\documentclass[runningheads]{llncs}

\usepackage[T1]{fontenc}
\usepackage{graphicx}
\usepackage{booktabs}
\usepackage{tabularx}
\usepackage{array}
\usepackage{amsmath}
\usepackage{url}
\usepackage{microtype}
\usepackage{enumitem}
\usepackage{float}
\usepackage[hidelinks]{hyperref}
\setlist{nosep,leftmargin=*}
\newcommand{\aw}{\texttt{article-\allowbreak{}writer}}
\newcommand{\awi}{\texttt{article-writer-\allowbreak improving}}
\newcommand{\pce}{\texttt{pdf-comparison-\allowbreak evaluation}}

\newcommand{\tasi}{Tasi Harness}
\newcommand{\geis}{GEIS}
\newcommand{\tba}{\texttt{tasi-browser-\allowbreak automation}}
\newcommand{\ad}{\texttt{architecture-\allowbreak diagram}}
\newcommand{\diff}{\Delta}
\newcolumntype{Y}{>{\raggedright\arraybackslash}X}

\begin{document}

\title{GEIS: A Generation--Evaluation--Improvement Loop of Agent Skills for Long-Form Article Generation}
\titlerunning{GEIS for Long-Form Article Generation}

\author{Jiale Zhang\inst{1} \and
Juntao Hu \inst{2} \and
Zhijian Ou\inst{1,2}\thanks{Corresponding author (ozj@tsinghua.edu.cn). This work is supported by the National Science and Technology Major Project (2023ZD0121401).}
}
\authorrunning{J. Zhang et al.}

\institute{
Speech Processing and Machine Intelligence (SPMI) Lab, Tsinghua University, China\and
TasiTech Co., Ltd., China
}

\maketitle

\begin{abstract}
Long-form article generation remains difficult for large language models because it combines long context, long instructions, and long outputs. Existing multi-agent pipelines such as STORM improve information coverage by simulating role-specialized agents, but their capabilities are often entangled in prompts and fixed procedures, making them hard to inspect, reuse, or iteratively improve. This paper presents \geis{} (Generation--Evaluation--Improvement loop of agent Skills), a loop of named and declarative skills for Wikipedia-style long-form article generation. Implemented and evaluated in \tasi{}, \geis{} composes skills for article writing, browser-based evidence and image collection, diagram rendering, PDF-aware pairwise evaluation, and rule-level skill improvement. Its core writing skill follows Request, Plan, Draft, Audit, Refine, and Deliver; the pairwise evaluation skill produces structured quality reports; and the improvement skill maps recurrent findings into permanent patches to the writing skill in our 20-topic experiment. We evaluate \geis{} on 20 Wikipedia Featured Article topics. Under the same generation backend, \geis{} improves over the Tasi Harness default writer by 8.0 points on a 100-point PDF quality rubric and outperforms STORM on the two comparable writing dimensions, structural quality and content quality. In the 20-topic improvement experiment, the patched writing skill raises the average score from 82.90 to 86.95, with 17 out of 20 topics improved and the gain mainly coming from content quality. These results show that long-form generation can be reframed from a fixed workflow into an inspectable, modular, and evaluation-guided improvement loop.
\end{abstract}

\section{Introduction}

Large language models (LLMs) have become strong generators of short answers, summaries, code snippets, and conversational responses~\cite{achiam2024gpt4}. However, many real writing tasks are not short-form generation problems. Wikipedia-style article writing, technical report preparation, proposal drafting, and analytical document creation require a system to collect information, organize it into a coherent structure, maintain factual consistency across sections, and produce a polished deliverable. We refer to such tasks as \emph{L3} long-form generation tasks because they jointly involve long context, long instructions, and long outputs, a combination emphasized in recent long-output and long-context generation studies~\cite{bai2024longwriter,wu2024longgenbench}.

The L3 setting exposes several weaknesses of single-pass LLM generation. First, long contexts can cause information loss and attention imbalance, including the well-known ``lost in the middle'' phenomenon~\cite{liu2024lost}. Second, long articles require several objectives to be optimized simultaneously: coverage, structure, coherence, factual reliability, style, and delivery readiness. A single autoregressive pass has no explicit mid-course quality gate. Third, long outputs make hallucination, repetition, and unsupported factual claims more likely, especially when generation must integrate many retrieved facts~\cite{huang2024hallucination,zhao2024ragcot,cao2025jsarag}. When a paragraph-level error appears early, it can be amplified by later sections or become hard to detect after the document has been assembled.

A natural solution is to decompose writing into multiple stages or multiple agents. STORM~\cite{shao2024storm}, for example, simulates Wikipedia writers and experts to explore a topic from multiple perspectives before drafting the article. PEER~\cite{schick2022peer} models writing as a plan-edit-explain-repeat process, while collaborative and multi-agent systems study how LLMs can coordinate through conversation~\cite{wu2023autogen,wu2025collabllm}. These systems show that decomposition helps. Yet, in many implementations, role boundaries remain implicit in prompts, tool use is entangled with the process, and evaluation does not naturally feed back into reusable writing rules.

This paper explores a different organization principle: \emph{agent skills}~\cite{anthropic2025skills}. Concrete skill names are typeset in monospace; see Appendix~\ref{app:skills} for the full list. In the skill paradigm, a reusable ability is described by a small metadata block and a skill document, optionally accompanied by scripts, templates, and resources. A model loads the skill only when the current task matches its description. Compared with a fixed multi-agent pipeline, this style makes capability boundaries explicit: one skill can focus on writing, another on browser retrieval, another on diagram generation, and another on evaluation. The goal is not merely to rename agents as skills, but to make long-form writing controllable, auditable, and open to iterative optimization.

\begin{figure}[t]
\centering
\includegraphics[width=0.95\textwidth]{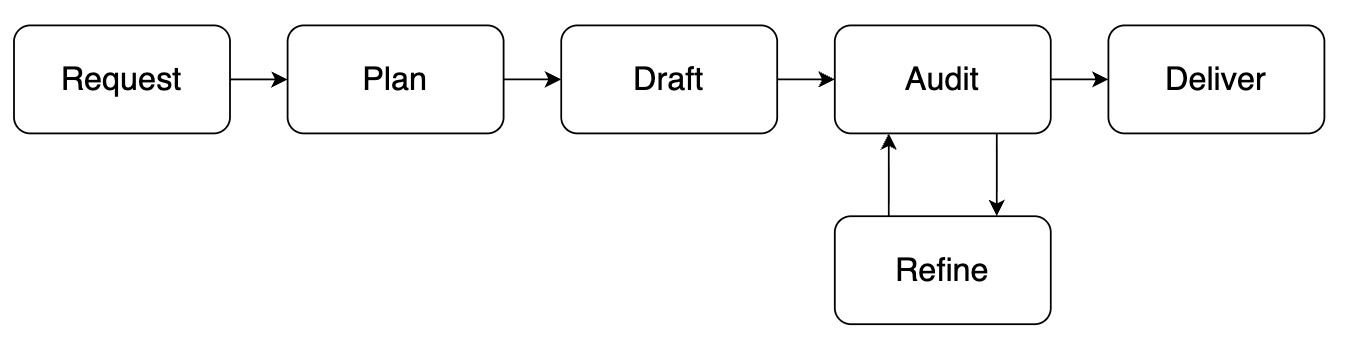}
\caption{The six-stage \aw{} process. Request normalizes the user task, Plan constructs the article outline, Draft writes the full article, Audit checks quality, Refine revises weak parts, and Deliver emits the final text or document.}
\label{fig:process}
\end{figure}

We present \geis{}, a Generation--Evaluation--Improvement loop of agent skills for professional long-form article generation. At the generation side, \geis{} centers on the \aw{} skill, which turns writing into six explicit stages: Request, Plan, Draft, Audit, Refine, and Deliver. At the tool side, browser automation, image search, and architecture diagramming are delegated to the \tba{} and \ad{} skills, respectively. At the evaluation side, \geis{} uses the \pce{} skill to produce symmetric PDF-aware comparison reports. Finally, the \awi{} skill converts recurrent evaluation feedback into permanent patches to the writing skill in the experiment reported here. Figure~\ref{fig:process} illustrates the core writing process.

Our contributions are threefold. First, we formulate a skill-based decomposition for long-form generation, separating writing, retrieval, image and diagram handling, evaluation, and improvement into auditable units. Second, we design an evaluation-and-improvement loop that maps pairwise quality reports back to writing rules, thereby turning external judgment into reusable skill evolution. Third, we evaluate the framework on 20 Wikipedia Featured Article topics and report both cross-system comparisons and a 20-topic improvement experiment. The improvement experiment shows an average gain of 4.05 points after patching the writing skill.

\section{Related Work}

\subsection{Long-Form Generation and Structural Control}

Long-form generation has recently attracted increasing attention because frontier models still struggle to generate very long, well-structured text. LongWriter~\cite{bai2024longwriter} approaches the problem from the training-data perspective by constructing long-output examples and improving the ability to produce 10,000-word responses. LongGenBench~\cite{wu2024longgenbench} evaluates long-form generation under temporal and spatial instruction constraints, distinguishing local instruction following from global structural completeness. These works highlight a central challenge: generating long text is not simply generating more tokens. A high-quality article must have a coherent global plan, stable local transitions, and coverage of topic-specific components.

Traditional automatic text metrics such as ROUGE~\cite{lin2004rouge} are insufficient for this setting because a generated article often has no single reference answer and may be valid under multiple outlines. Factuality metrics such as FActScore~\cite{min2023factscore} are more appropriate for factual precision but still do not cover layout, hierarchy, visual coherence, and document delivery. Consequently, LLM-as-a-judge methods~\cite{zheng2023judge} are increasingly used for structured quality assessment, although they require careful rubric design and bias control.

\subsection{Agentic and Multi-Agent Writing Pipelines}

LLM-based agents use language models not only to generate text but also to plan, call tools, inspect observations, and revise outputs~\cite{xi2025agents}. ReAct~\cite{yao2023react} interleaves reasoning and acting, Toolformer~\cite{schick2023toolformer} demonstrates tool-use learning, and retrieval-augmented generation (RAG)~\cite{lewis2020rag,zhao2024ragcot,cao2025jsarag} grounds generation in external evidence. For collaborative generation, AutoGen~\cite{wu2023autogen} and related frameworks organize multiple LLM instances into conversational teams.

STORM~\cite{shao2024storm} is the closest prior system to our setting. It builds Wikipedia-like articles by first surveying related articles, identifying multiple perspectives, simulating dialogues between Wikipedia writers and experts, generating an outline, and then drafting sections with retrieved references. STORM improves coverage, but it also exposes design limitations that motivate \geis{}. Its roles and process are tightly coupled in prompt templates and DSPy modules~\cite{khattab2024dspy}; intermediate quality gates are weak; and retrieval, filtering, and writing are deeply bound to the pipeline. In contrast, \geis{} makes each capability an independent skill with declarative scope, and it adds explicit audit and improvement stages.

\subsection{Skills and Self-Improving Agents}

Agent skills provide a modular abstraction for packaging task-specific instructions, code, and resources~\cite{anthropic2025skills}. The key idea is progressive disclosure: only relevant skills are loaded into the model context. This makes skills attractive for L3 generation, where context budget is scarce and irrelevant instructions can interfere with writing.

Self-improvement has also been explored in agent systems. Reflexion~\cite{shinn2023reflexion} uses verbal feedback to improve future behavior. SkillRL~\cite{xia2026skillrl} and Skill-R1~\cite{vishe2026skillr1} investigate reinforcement-learning-style skill evolution. Our work is complementary: instead of training the underlying model, we optimize a skill document through evaluation-driven rule updates. This is a lightweight, engineering-oriented form of process improvement suitable for professional writing systems where changing the model is expensive or impossible.

\section{GEIS: Skill-Based Long-Form Article Generation}

\subsection{Design Goals}

\geis{} is designed around four goals. \emph{Explicit capability boundaries} require that writing, retrieval, image collection, diagram rendering, evaluation, and improvement be separable. \emph{Quality gates} require that drafts be checked before delivery, rather than relying on a single final pass. \emph{Traceable evaluation} requires that scoring reports explain which dimensions were weak and why. \emph{Iterative skill optimization} requires that repeated weaknesses can be translated into future writing rules.

The overall framework is implemented as a collection of named skills used by \tasi{}. Each skill is represented by a \texttt{SKILL.md} file with frontmatter fields such as \texttt{name}, \texttt{description}, \texttt{category}, and \texttt{license}. The \texttt{description} field is used for progressive disclosure. For example, the \aw{} skill declares that it generates and rewrites high-quality article text before Word or PDF formatting, and that complex tasks should follow Request $\rightarrow$ Plan $\rightarrow$ Draft $\rightarrow$ Audit $\rightarrow$ Refine $\rightarrow$ Deliver.

\subsection{The Core Writing Skill}

The \aw{} skill is the central writing component. Unlike a role-based multi-agent pipeline, it is organized by semantic stages rather than by agent names. The stages have the following responsibilities.

\textbf{Request.} The system restates the objective, audience, output type, and constraints. Missing information is handled by explicit assumptions rather than by shallow drafting.

\textbf{Plan.} The system builds a section-level outline, i.e., an outline at the granularity of article sections and subsections. The plan includes not only headings, but also expected depth, subtopics, and potential visual anchors. A visual anchor is a deterministic placeholder that reserves a later image or diagram insertion point for downstream rendering. This stage is a structural contract for the later draft.

\textbf{Draft.} The system writes a complete article, not a bullet-only skeleton. For non-brief requests, major sections are expected to contain substantive prose, examples, and domain-appropriate terminology.

\textbf{Audit.} The system checks structure, factual consistency, clarity, actionability, and audience fit. This stage directly addresses the failure mode of one-way pipelines that pass flawed intermediate output downstream.

\textbf{Refine.} Weak or thin sections are revised, redundant parts are removed, and missing evidence or examples are added.

\textbf{Deliver.} Only the final polished content is emitted to the user or written to a file. Intermediate notes remain internal. This output policy avoids leaking planning traces into the final document.

The skill further defines quality defaults. When the user does not ask for brevity, it must produce a substantive draft; each major section should contain developed paragraphs; and the final output must follow the planned structure. These rules turn vague writing preferences into inspectable constraints.

\subsection{Delegated Retrieval, Image, and Diagram Skills}

The writing skill delegates web evidence collection and real-world image search to the \tba{} skill and diagramming to the \ad{} skill. The browser skill uses a real browser controlled through the Chrome DevTools Protocol (CDP). Its operations include opening pages, waiting for page state, extracting semantic snapshots, interacting with elements, and extracting page content. The writing skill is therefore not responsible for low-level web navigation. It only requests evidence, image candidates, and traceable source information.

The \ad{} skill turns structured component-relation descriptions into HTML/SVG diagrams and PNG exports. The \tba{} skill supplies source-grounded real-world images. Instead of embedding images directly in the text, \aw{} uses deterministic anchors:
\begin{itemize}
  \item \texttt{[[DIAGRAM:id|title=...]]} for generated diagrams;
  \item \texttt{[[IMAGE:id|query=...|title=...]]} for real-world images retrieved through browser search.
\end{itemize}
Together, these diagram and image anchors decouple text generation from downstream rendering. A Markdown, Word, or PDF pipeline can later resolve the anchors into figures. The skill also requires anchor IDs to be unique and avoids visually duplicate figures.

\subsection{Comparison with STORM}

Table~\ref{tab:storm_skill} summarizes the main architectural differences between STORM and \geis{}. The key difference is not the number of LLM calls, but the unit of abstraction. STORM uses role prompts in a fixed pipeline. \geis{} uses declared skills whose scope can be inspected and changed independently.

\begin{table}[t]
\caption{Key differences between STORM and \geis{}.}
\label{tab:storm_skill}
\centering
\small
\begin{tabularx}{\textwidth}{p{0.22\textwidth}YY}
\toprule
Aspect & STORM-style pipeline & GEIS skill composition \\
\midrule
Capability boundary & Roles such as writer, expert, and outliner are embedded in prompts and pipeline modules. & Writing, retrieval, image handling, diagramming, evaluation, and improvement are declared as separate skills. \\
Context loading & Prompt resources are largely static once the pipeline is configured. & Progressive disclosure loads only task-relevant skills. \\
Quality control & Mainly prompt-level instructions and downstream merging. & Explicit Audit/Refine stages with quality checklists. \\
Tool integration & Retrieval, filtering, and generation are tightly coupled. & External tools are delegated to independent skills. \\
Visual delivery & Primarily text article generation. & Image and diagram anchors support visual rendering and PDF delivery. \\
Improvement & No native evaluation-to-rule update loop. & Evaluation reports can become improvement plans and permanent patches; session overrides are supported by the improving skill but are not the main setting in our 20-topic experiment. \\
\bottomrule
\end{tabularx}
\end{table}

\section{Evaluation and Improvement Skills}

\subsection{PDF Quality Evaluation}

The experiments use the \pce{} skill as the evaluation skill. It applies the PDF quality rubric but performs symmetric comparison between two peer documents rather than treating one document as a gold standard. The rubric uses four weighted dimensions for the first two experiments: structural quality (35 points), content quality (35 points), visual quality (15 points), and PDF delivery quality (15 points). Structural quality includes hierarchy, coherence, formatting consistency, and required components. Content quality includes task alignment, correctness, depth, and profile-specific requirements. Visual quality measures image relevance, legibility, data consistency, and caption consistency. PDF delivery quality measures searchable text, font stability, navigation metadata, page geometry, and accessibility signals.

Pairwise comparison is used because absolute single-document scores can drift across calls, and many quality differences are inherently comparative: two articles on the same topic may both be internally consistent but differ in depth, structure, or coverage. The \pce{} skill applies the same rubric to both documents and then explains dimension-level score differences and topic-level gaps. The comparison skill enforces symmetry: neither document is treated as a gold standard, and path order should not influence score strictness.

The comparison report contains topic alignment, scorecards, arithmetic verification, findings, and a revision roadmap. These fields make the evaluation output suitable for downstream improvement. The report is not only a score; it is an executable diagnosis of what should be revised.

\subsection{From Evaluation to Skill Improvement}

The \awi{} skill closes the loop. It reads an evaluation report, determines which findings are within the control of \aw{}, maps them to specific writing-rule targets, and produces an improvement plan. It distinguishes authoring issues from extraction artifacts, PDF-export issues, and benchmark-only differences. For example, thin sections, missing citations, poor transitions, or missing conclusion sections are authoring issues. Missing bookmarks or font embedding problems belong to export tooling and should not be patched into writing rules.

\begin{figure}[t]
\centering
\includegraphics[width=0.92\textwidth]{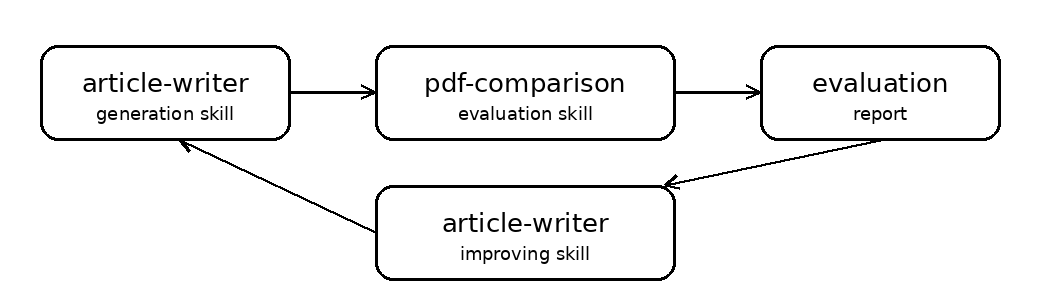}
\caption{Closed-loop optimization. The writing skill generates an article, the pairwise evaluation skill produces a report, and the improvement skill converts recurrent report findings into permanent writing-skill patches in the reported experiment.}
\label{fig:loop}
\end{figure}

The improvement skill can emit three output levels. A \emph{plan} explains the root causes and proposed fixes; a \emph{session override} gives temporary constraints for a next generation run; and a \emph{permanent patch} modifies the skill itself when a problem recurs across reports or represents a blocking or major finding (P0/P1 in the evaluation report). In the 20-topic experiment reported in this paper, the measured improvement comes from permanent patches to \aw{}, not from session-only constraints. Figure~\ref{fig:loop} shows the generate--evaluate--improve loop.

In the 20-topic improvement run, recurrent findings produced eight improvement items: minimum source requirements, inline citations for factual claims, stable image sources, deeper section budgets, bridging transitions, formal bibliography formatting, topic completeness checks, and mandatory conclusions. These patches do not change the underlying LLM. They change the process that the LLM follows.

\section{Experimental Setup}

\subsection{Implementation Platform: Tasi Harness}

All experiments are executed in \tasi{}, which serves as the agent runtime rather than the proposed method itself. According to the official repository, Tasi Harness is a local-first desktop AI agent built with Electron, TypeScript, React, and Vite; it integrates a streaming tool-calling runtime, multimodal attachments, browser automation, persistent memory, editable skills, scheduled tasks, document writing/export, and a personal knowledge base~\cite{tasiharness2026}. Its skill system reads, creates, patches, uploads, optimizes, and installs \texttt{SKILL.md} assets, while its browser automation module supports CDP-backed Chromium navigation, semantic snapshots with \texttt{@e} references, extraction, screenshots, and PDF capture~\cite{tasiharness2026}.

In this work, \tasi{} provides four implementation services: the iterative agent loop and prompt assembly, workspace-scoped file operations, skill loading and patching, and document export. \geis{} is therefore the skill composition deployed on this runtime: \aw{} produces the article, \tba{} collects web evidence and images, \ad{} renders diagrams, \pce{} evaluates documents through symmetric comparison, and \awi{} converts evaluation findings into writing-skill updates.

\subsection{Dataset}

We build an evaluation set from 20 English Wikipedia Featured Articles. Featured Articles are community-reviewed and generally strong in neutrality, comprehensiveness, source reliability, and writing style. We use them in three ways: as topic names for generation, as high-quality examples for gap diagnosis, and as external reference points for article quality. In system-to-system comparisons, the Wikipedia PDFs are not used as gold outputs; in the improvement diagnosis, they are compared symmetrically and serve only as high-quality reference documents rather than authoritative answers.

Table~\ref{tab:dataset} lists the topics. The set covers history and culture, software engineering, computer science, artificial intelligence, networking, design, and biography. This diversity prevents conclusions from depending on a single technical domain.

\begin{table}[H]
\caption{The 20-topic Wikipedia Featured Article evaluation set.}
\label{tab:dataset}
\centering
\scriptsize
\begin{tabularx}{\textwidth}{rYl rYl}
\toprule
\# & Topic & Category & \# & Topic & Category \\
\midrule
1 & 1689 Boston revolt & History & 11 & Great Wall of China & History/culture \\
2 & 2008 World Science Festival & Science/culture & 12 & iMac G3 & Technology/design \\
3 & Autonomous agent & AI & 13 & Internet protocol suite & Networking \\
4 & Business Process Model and Notation & Software engineering & 14 & Knowledge graph & AI \\
5 & Compiler & Computer science & 15 & Manchester Baby & Computing history \\
6 & Data-flow diagram & Software engineering & 16 & Microservices & Software architecture \\
7 & Distributed computing & Computer science & 17 & Neural network & Machine learning \\
8 & Edge computing & Computer science & 18 & Prompt engineering & AI \\
9 & Finite-state machine & Computer science & 19 & Taylor Swift & Biography \\
10 & Freedom Monument & Architecture/history & 20 & Unified Modeling Language & Software engineering \\
\bottomrule
\end{tabularx}
\end{table}

\subsection{Models, Baselines, and Evaluation Profiles}

The generation and evaluation chains are separated. All generation-side calls use GPT-5.4 through \tasi{} for the harness default writer, \aw{}, and auxiliary skills. Evaluation calls use Qwen 3.5 Plus through \texttt{qwen-code}. Separating the generator and judge reduces self-evaluation bias and makes the comparison less dependent on the writing style of a single model.

We conduct three experiments: (i) \tasi{} default writing vs. \geis{}, (ii) STORM vs. \geis{}, and (iii) pre-improving vs. post-improving \geis{} on 20 topics. We compare three writing paths. \textbf{Harness default} writes directly without loading \aw{} or auxiliary skills. \textbf{GEIS} loads \aw{} plus retrieval, image, and diagramming skills. \textbf{STORM} runs the STORM pre-writing, outlining, and section-writing pipeline with the same topic names.

The first two experiments use the full PDF profile: structural quality (A, 35), content quality (B, 35), visual quality (C, 15), and PDF delivery quality (D, 15). The 20-topic improvement experiment uses a writing-focused profile with structure and content only (50+50). It evaluates Markdown/article content and deliberately removes image insertion and PDF-export dimensions; the goal is to isolate whether the patched writing skill improves authoring quality.

\section{Results}

\subsection{GEIS vs. Harness Default Writing}

Table~\ref{tab:default} reports the comparison between harness default writing and \geis{} on all 20 topics. \geis{} improves the average total score from 74.1 to 82.1, a gain of 8.0 points. The improvement appears in structure (+2.6), content (+3.2), and visual quality (+2.2), while PDF delivery remains equal. This pattern indicates that the gain is caused by the writing process and visual-anchor process, not by a better PDF exporter.

\begin{table}[t]
\caption{Tasi Harness default writing vs. \geis{} on 20 topics. Scores use the 100-point PDF profile.}
\label{tab:default}
\centering
\small
\begin{tabular}{lrrrr}
\toprule
Dimension & Harness & GEIS & $\diff$ & Max. \\
\midrule
A. Structural quality & 26.0 & 28.6 & +2.6 & 35 \\
B. Content quality & 27.8 & 31.0 & +3.2 & 35 \\
C. Visual quality & 10.0 & 12.2 & +2.2 & 15 \\
D. PDF delivery quality & 10.3 & 10.3 & 0.0 & 15 \\
\midrule
Total & 74.1 & 82.1 & +8.0 & 100 \\
\bottomrule
\end{tabular}
\end{table}

Qualitatively, the evaluation reports repeatedly attribute the difference to clearer section hierarchy, more complete topic coverage, and better use of images or diagrams. Harness default writing sometimes produces readable prose, but it more often omits practice-oriented sections or loses figures during conversion. \geis{}'s Plan and Audit stages make the outline more stable, and the diagram/image anchors provide a systematic way to include visuals.

\subsection{\geis{} vs. STORM}

Table~\ref{tab:storm_result} compares \geis{} with STORM. Only the two writing dimensions, A and B, are directly comparable, because STORM outputs are text-centered and are not processed by \geis{}'s visual-anchor and PDF-delivery pipeline. \geis{} leads STORM by 5.6 points in structure and 2.2 points in content. The larger structural gap supports our design argument: explicit Plan/Deliver constraints and Audit/Refine gates help maintain section boundaries and complete article components.

\begin{table}[t]
\caption{\geis{} vs. STORM. C and D are not scored for STORM because the STORM output is not processed by \geis{}'s visual-anchor and PDF-delivery pipeline.}
\label{tab:storm_result}
\centering
\small
\begin{tabular}{lrrrr}
\toprule
Dimension & STORM & GEIS & $\diff$ & Max. \\
\midrule
A. Structural quality & 22.6 & 28.2 & +5.6 & 35 \\
B. Content quality & 28.4 & 30.6 & +2.2 & 35 \\
C. Visual quality & N/A & 12.0 & -- & 15 \\
D. PDF delivery quality & N/A & 10.0 & -- & 15 \\
\midrule
Comparable total (A+B) & 51.0 & 58.8 & +7.8 & 70 \\
Full GEIS total & -- & 80.8 & -- & 100 \\
\bottomrule
\end{tabular}
\end{table}

STORM remains competitive on some content-depth dimensions, especially when its multi-perspective exploration identifies rich historical or technical angles. However, its fixed pipeline does not naturally enforce a final conclusion, visual hierarchy, or diagram anchoring. \geis{}'s advantage is therefore not simply stronger retrieval; it is a more controllable writing process.

\subsection{20-Topic Improvement Experiment}

We report the 20-topic improvement result from the code package. In this experiment, \awi{} analyzes 20 pairwise evaluation reports, applies permanent patches to the writing skill, regenerates all 20 articles, and then compares each post-improving article against its corresponding pre-improving article.

Table~\ref{tab:improve_aggregate} gives the aggregate result. The average score increases from 82.90 to 86.95, a gain of 4.05 points. Content quality accounts for almost all of the improvement (+3.85), while structure changes only slightly (+0.20). This is consistent with the actual patches, which mainly enforce source collection, inline citations, deeper technical coverage, formal bibliography, and topic completeness.

\begin{table}[H]
\caption{20-topic improvement result. The profile uses A. structure (50) and B. content (50).}
\label{tab:improve_aggregate}
\centering
\small
\begin{tabular}{lrrrr}
\toprule
Dimension & Pre-improving & Post-improving & $\diff$ & Max. \\
\midrule
A. Structural quality & 43.45 & 43.65 & +0.20 & 50 \\
B. Content quality & 39.45 & 43.30 & +3.85 & 50 \\
\midrule
Total & 82.90 & 86.95 & +4.05 & 100 \\
\bottomrule
\end{tabular}
\end{table}

Table~\ref{tab:improve_topic} shows the per-topic deltas, with columns ordered as pre-improving, post-improving, and post-minus-pre change. The patched skill improves 17 of 20 topics. The largest gains are observed on 2008 World Science Festival (+11), Freedom Monument (+9), Great Wall of China (+9), and Internet protocol suite (+9). Three topics decline: Autonomous agent (-5), Knowledge graph (-2), and Microservices (-2). The declines are instructive because they reveal a limitation of uniform skill patches. For example, Autonomous agent loses academic depth and reference count compared with its strong original version, while Microservices loses some H3-level practical structure. Skill evolution therefore needs topic-sensitive policies, not only global rules.

\begin{table}[H]
\caption{Per-topic results for the 20-topic improvement experiment. Pre and Post denote pre-improving and post-improving articles, respectively.}
\label{tab:improve_topic}
\centering
\scriptsize
\begin{tabularx}{\textwidth}{Yrrrr|Yrrrr}
\toprule
Topic & Pre & Post & $\diff$ & $\diff_B$ & Topic & Pre & Post & $\diff$ & $\diff_B$ \\
\midrule
2008 World Science Festival & 79 & 90 & +11 & +6 & Compiler & 83 & 87 & +4 & +5 \\
Freedom Monument & 79 & 88 & +9 & +5 & Prompt engineering & 83 & 87 & +4 & +2 \\
Great Wall of China & 81 & 90 & +9 & +6 & Edge computing & 84 & 87 & +3 & +5 \\
Internet protocol suite & 80 & 89 & +9 & +8 & Taylor Swift & 82 & 85 & +3 & +2 \\
Manchester Baby & 81 & 88 & +7 & +4 & UML & 85 & 88 & +3 & +4 \\
1689 Boston Revolt & 82 & 88 & +6 & +4 & Neural network & 84 & 86 & +2 & +3 \\
Distributed computing & 82 & 88 & +6 & +7 & BPMN & 84 & 85 & +1 & +2 \\
Finite-state machine & 83 & 89 & +6 & +7 & Data-flow diagram & 85 & 86 & +1 & +2 \\
iMac G3 & 82 & 88 & +6 & +5 & Knowledge graph & 87 & 85 & -2 & +2 \\
Microservices & 86 & 84 & -2 & +1 & Autonomous agent & 86 & 81 & -5 & -3 \\
\bottomrule
\end{tabularx}
\vspace{0.3em}
\footnotesize{$\diff$ is post-minus-pre total-score change. $\diff_B$ is post-minus-pre content-dimension change.}
\end{table}

\section{Analysis}

The results suggest that agent skills help long-form generation by separating responsibilities that are usually entangled in long prompts. The writing skill does not carry browser-operation, image-search, or diagram-rendering instructions; it only delegates to \tba{} or \ad{} when evidence or visuals are required. This preserves context for article planning and makes each capability independently inspectable. The six-stage writing process also turns a vague writing goal into checkable obligations, so thin sections, missing citations, unresolved anchors, and absent conclusions can be detected before delivery.

The 20-topic improvement experiment shows that most gains come from content rather than structure. The patches mainly increase source requirements, inline citations, formal bibliography formatting, section-depth budgets, topic-completeness checks, and conclusion synthesis. These changes directly improve factual support and coverage. Structural gains are weaker because structure is topic-sensitive: a global rule that adds subsections can help thin articles but may harm an already effective outline. This explains the regressions on Autonomous agent, Knowledge graph, and Microservices, and motivates topic-adaptive patch policies.

Pairwise evaluation is central to the loop. A single score can mark an article as weak, but it rarely tells the system which reusable writing rule should change. The pairwise reports identify shared-topic depth gaps, missing profile components, citation weaknesses, and factual contradictions while filtering out PDF-export defects and benchmark-only topics. Thus \awi{} is not merely prompt polishing; it is a controlled translation from evaluation evidence to authoring rules.

\section{Limitations and Future Work}

This work still depends on LLM-as-a-judge. Structured rubrics, arithmetic verification, and pairwise symmetry reduce common errors, but judge variance and scale drift remain; future work should add expert review and more fine-grained factual checks. The reference PDFs are high-quality examples rather than gold answers, which is appropriate for open-ended article generation but means the scores should be read as quality indicators, not absolute truth.

The current improvement method is rule-based. It is practical and auditable because it patches \texttt{SKILL.md} rather than model weights, but uniform rules can fail on topics that need special structure or academic depth. Future work can learn topic-conditioned skill variants, combine patching with reinforcement-learning-style skill evolution~\cite{xia2026skillrl,vishe2026skillr1}, and extend \geis{} to professional documents that require stricter citation styles, domain risk control, tables from structured data, or human-in-the-loop review.

\section{Conclusion}

We presented \geis{}, a Generation--Evaluation--Improvement loop of agent skills for long-form article generation. \geis{} decomposes writing, retrieval, image collection, diagramming, evaluation, comparison, and improvement into named and declarative skills deployed in \tasi{}. Experiments on 20 Wikipedia Featured Article topics show that \geis{} improves over the Tasi Harness default writer by 8.0 points and outperforms STORM on the comparable writing dimensions of structural and content quality. The 20-topic improvement experiment further shows that skill patching raises the average score from 82.90 to 86.95 and improves 17 out of 20 topics. These results support a broader conclusion: for long-form generation, the structure and inspectability of the skill loop can be as important as the capability of the underlying model.

\appendix
\renewcommand{\theHsection}{appendix.\Alph{section}}
\section{Skill Specifications Used in GEIS}
\label{app:skills}

Table~\ref{tab:skill_specs} summarizes the skill specifications used in the experiments. Full Markdown assets will be open-sourced upon acceptance of this work.

\begin{table}[H]
\caption{Skill specifications used by \geis{}.}
\label{tab:skill_specs}
\centering
\scriptsize
\begin{tabularx}{\textwidth}{p{0.20\textwidth}p{0.20\textwidth}Y}
\toprule
Skill & Role & Specification excerpt \\
\midrule
\aw{} & Article generation & Request $\rightarrow$ Plan $\rightarrow$ Draft $\rightarrow$ Audit $\rightarrow$ Refine $\rightarrow$ Deliver; improved rules require sources, inline citations, bibliography, stable images, depth budgets, transitions, topic-completeness checks, and conclusions. \\
\tba{} & Web evidence & Uses browser open/state/snapshot/wait/find/extract operations with \texttt{@e} semantic references; opened pages and source URLs support factual claims and image provenance. \\
\ad{} & Diagram rendering & Converts component--relation descriptions into standalone HTML/SVG and PNG; \aw{} calls it through \texttt{[[DIAGRAM:...]]} anchors. \\
\pce{} & Pairwise scoring & Applies the rubric symmetrically to two peer documents, aligns shared topics, checks contradictions, and explains dimension-level and topic-level score differences without treating either document as gold. \\
\awi{} & Skill improvement & Filters evaluation reports for authoring-fixable gaps, maps recurrent findings to \aw{} rules, and emits an improvement plan and, for the reported experiment, permanent patches. \\
\bottomrule
\end{tabularx}
\end{table}


\begin{thebibliography}{99}
\bibitem{achiam2024gpt4}
OpenAI, Achiam, J., Adler, S., Agarwal, S., Ahmad, L., Akkaya, I., et al.: GPT-4 technical report. arXiv preprint arXiv:2303.08774 (2024)

\bibitem{liu2024lost}
Liu, N.F., Lin, K., Hewitt, J., Paranjape, A., Bevilacqua, M., Petroni, F., Liang, P.: Lost in the middle: How language models use long contexts. Transactions of the Association for Computational Linguistics 12, 157--173 (2024)

\bibitem{huang2024hallucination}
Huang, L., Yu, W., Ma, W., Zhong, W., Feng, Z., Wang, H., Chen, Q., Peng, W., Feng, X., Qin, B., Liu, T.: A survey on hallucination in large language models: Principles, taxonomy, challenges, and open questions. ACM Transactions on Information Systems 43(2) (2024). \doi{10.1145/3703155}

\bibitem{shao2024storm}
Shao, Y., Jiang, Y., Kanell, T., Xu, P., Khattab, O., Lam, M.: Assisting in writing Wikipedia-like articles from scratch with large language models. In: Proceedings of NAACL-HLT 2024, pp. 6252--6278 (2024). \doi{10.18653/v1/2024.naacl-long.347}

\bibitem{schick2022peer}
Schick, T., Dwivedi-Yu, J., Jiang, Z., Petroni, F., Lewis, P., Izacard, G., You, Q., Nalmpantis, C., Grave, E., Riedel, S.: PEER: A collaborative language model. arXiv preprint arXiv:2208.11663 (2022)

\bibitem{wu2023autogen}
Wu, Q., Bansal, G., Zhang, J., Wu, Y., Li, B., Zhu, E., Jiang, L., Zhang, X., Zhang, S., Liu, J., Awadallah, A.H., White, R.W., Burger, D., Wang, C.: AutoGen: Enabling next-gen LLM applications via multi-agent conversation. arXiv preprint arXiv:2308.08155 (2023)

\bibitem{wu2025collabllm}
Wu, S., Galley, M., Peng, B., Cheng, H., Li, G., Zhu, Y., Leskovec, J., Gao, J.: CollabLLM: From passive responders to active collaborators. arXiv preprint arXiv:2502.00640 (2025)

\bibitem{bai2024longwriter}
Bai, Y., Zhang, J., Lv, X., Zheng, L., Zhu, S., Hou, L., Dong, Y., Tang, J., Li, J.: LongWriter: Unleashing 10,000+ word generation from long context LLMs. arXiv preprint arXiv:2408.07055 (2024)

\bibitem{wu2024longgenbench}
Wu, Y., Hee, M.S., Hu, Z., Lee, R.K.-W.: LongGenBench: Benchmarking long-form generation in long context LLMs. arXiv preprint arXiv:2409.02076 (2024)

\bibitem{lin2004rouge}
Lin, C.-Y.: ROUGE: A package for automatic evaluation of summaries. In: Text Summarization Branches Out, pp. 74--81 (2004)

\bibitem{min2023factscore}
Min, S., Krishna, K., Lyu, X., Lewis, M., Yih, W.-t., Koh, P.W., Iyyer, M., Zettlemoyer, L., Hajishirzi, H.: FActScore: Fine-grained atomic evaluation of factual precision in long form text generation. In: Proceedings of EMNLP 2023, pp. 12076--12100 (2023). \doi{10.18653/v1/2023.emnlp-main.741}

\bibitem{zheng2023judge}
Zheng, L., Chiang, W.-L., Sheng, Y., Zhuang, S., Wu, Z., Zhuang, Y., Lin, Z., Li, Z., Li, D., Xing, E., Zhang, H., Gonzalez, J.E., Stoica, I.: Judging LLM-as-a-judge with MT-Bench and Chatbot Arena. In: Advances in Neural Information Processing Systems (2023)

\bibitem{xi2025agents}
Xi, Z., Chen, W., Guo, X., He, W., Ding, Y., Hong, B., Zhang, M., Wang, J., Jin, S., Zhou, E., et al.: The rise and potential of large language model based agents: A survey. Science China Information Sciences 68(2) (2025). \doi{10.1007/s11432-024-4222-0}

\bibitem{yao2023react}
Yao, S., Zhao, J., Yu, D., Du, N., Shafran, I., Narasimhan, K., Cao, Y.: ReAct: Synergizing reasoning and acting in language models. In: International Conference on Learning Representations (2023)

\bibitem{schick2023toolformer}
Schick, T., Dwivedi-Yu, J., Dessi, R., Raileanu, R., Lomeli, M., Zettlemoyer, L., Cancedda, N., Scialom, T.: Toolformer: Language models can teach themselves to use tools. arXiv preprint arXiv:2302.04761 (2023)

\bibitem{lewis2020rag}
Lewis, P., Perez, E., Piktus, A., Petroni, F., Karpukhin, V., Goyal, N., Kuttler, H., Lewis, M., Yih, W.-t., Rocktaschel, T., Riedel, S., Kiela, D.: Retrieval-augmented generation for knowledge-intensive NLP tasks. In: Advances in Neural Information Processing Systems (2020)

\bibitem{zhao2024ragcot}
Zhao, Y., Cao, H., Zhao, X., Ou, Z.: An empirical study of retrieval augmented generation with chain-of-thought. In: Proceedings of ISCSLP (2024)

\bibitem{cao2025jsarag}
Cao, H., Wu, Y., Cai, Y., Zhao, X., Ou, Z.: Improving end-to-end training of retrieval-augmented generation models via joint stochastic approximation. arXiv preprint arXiv:2508.18168 (2025)


\bibitem{tasiharness2026}
TasiTech: Tasi Harness. GitHub repository. \url{https://github.com/TasiTech/tasi-harness} (2026), last accessed 4 July 2026

\bibitem{anthropic2025skills}
Anthropic: Agent Skills. \url{https://www.anthropic.com/news/agent-skills} (2025), last accessed 28 May 2026

\bibitem{khattab2024dspy}
Khattab, O., Singhvi, A., Maheshwari, P., Zhang, Z., Santhanam, K., et al.: DSPy: Compiling declarative language model calls into state-of-the-art pipelines. In: International Conference on Learning Representations (2024)

\bibitem{shinn2023reflexion}
Shinn, N., Cassano, F., Berman, E., Gopinath, A., Narasimhan, K., Yao, S.: Reflexion: Language agents with verbal reinforcement learning. In: Advances in Neural Information Processing Systems (2023)

\bibitem{xia2026skillrl}
Xia, P., Chen, J., Wang, H., Liu, J., Zeng, K., Wang, Y., Han, S., Zhou, Y., Zhao, X., Chen, H., Zheng, Z., Xie, C., Yao, H.: SkillRL: Evolving agents via recursive skill-augmented reinforcement learning. arXiv preprint arXiv:2602.08234 (2026)

\bibitem{vishe2026skillr1}
Vishe, Y., Surana, R., Jiang, X., Huang, Z., Li, X., Kuang, N.L., Yu, T., Rossi, R.A., Shang, J., McAuley, J., Wu, J.: Skill-R1: Agent skill evolution via reinforcement learning. arXiv preprint arXiv:2605.09359 (2026)
\end{thebibliography}
\end{document}